\documentclass{article}

\PassOptionsToPackage{numbers,compress}{natbib}
\usepackage[preprint]{neurips_2024}

\usepackage[utf8]{inputenc}
\usepackage[T1]{fontenc}
\usepackage{hyperref}
\usepackage{url}
\usepackage{booktabs}
\usepackage{placeins}
\usepackage{amsfonts}
\usepackage{amsmath}
\usepackage{nicefrac}
\usepackage{microtype}
\usepackage{xcolor}

\title{The Era by Eon Benchmark: A Generated Enterprise Estate with Exact Ground
  Truth for Benchmarking LLM Agents}

\author{%
  \normalfont\small
  Benjamin Gruenbaum, Doron Porat, Assaf Natanzon\\
  Roy Zavida, Chen Dinachi, Or Itzahary\\[5pt]
  Eon\\
  \texttt{assaf@eon.io}
}

\begin{document}

\maketitle

\begin{abstract}
LLM agents for enterprise systems of record cannot be evaluated on
customer production data, and no existing substitute provides ground
truth. We present the Era by Eon Benchmark for evaluating LLM agents
that use enterprise tools. The benchmark is built around a complete
fictional company. It includes product simulators, company-specific
internal databases, benchmark questions, and computed answer keys.
Industry, company size, business model, application portfolio, and a
seed define each company. One seeded entity graph supplies shared
company data to simulators of Salesforce, Zendesk, Slack, Gong, and
other products. A question-conditioned generator creates the schemas
and records for internal databases. It takes shared entities, keys,
and values from the same graph before generating database-specific
facts. Both mechanisms therefore describe one consistent enterprise
estate. Every expected answer is computed from the final records, so
grading is exact. Design and answer-key checks validate the internal
databases. A realism scorecard and adversarial detector validate the
entity graph. Across 23 generated companies, the mean realism score
rose from 61.8 to 97.0, with zero records flagged as synthetic. In the
reported simulator-track comparison, nine models answered the same 33
questions three times each. Accuracy estimates ranged from 42.4\% to
76.8\%, and three of 36 pairwise differences remained supported after
correction.
\end{abstract}

\section{Introduction}

Developers of LLM agents that operate over enterprise systems of
record, such as CRM, support, billing, communication, and document
platforms, face a fundamental data-access problem. An agent must be
developed and evaluated against data of the kind it will encounter in
production, yet such data belongs to customers and is confidential.
The available substitutes are inadequate: vendor sandboxes contain
little data and no history; manually constructed test environments
cover one or two systems and are costly to extend; and neither
provides ground truth, so a developer can observe an agent's behavior
but cannot measure its accuracy.

Prior benchmarks address parts of this problem with hand-built
environments. $\tau$-bench~\citep{yao2024tau} constructs two small
domains (retail and airline) with manually authored databases and
tools. CRMArena~\citep{huang2025crmarena} and
CRMArena-Pro~\citep{huang2025crmarenapro} populate a single Salesforce
organization from a hand-designed schema.
TheAgentCompany~\citep{xu2024theagentcompany} instantiates a fictional
software firm across four self-hosted collaboration tools with
manually authored tasks. WorkArena~\citep{drouin2024workarena} and
WorkArena++~\citep{boisvert2024workarenapp} evaluate browser agents on
a standard ServiceNow installation. These benchmarks are carefully
engineered, and the present work builds on several of their design
decisions. However, each covers at most a few systems, each task set
is manually authored or curated, extension requires substantial
engineering effort, and the realism of the underlying data is asserted
rather than measured.

This paper presents the Era by Eon Benchmark (hereafter the benchmark).
It generates the evaluation environment in its entirety. Given four
parameters, namely industry, company size, business model, and the set
of business applications the company operates, together with a single
random seed, the benchmark deterministically generates the shared
entity graph of a complete fictional enterprise. The graph includes a
workforce with an organizational hierarchy and employment history, a
customer base, a sales pipeline, support tickets, recorded calls, chat
messages, and documents. The same seed reproduces the identical
dataset, which makes every evaluation exactly repeatable.

The generated company is served through product simulators. Each
simulator implements the interface of a real product (Salesforce,
Zendesk, Slack, Gong, S3, Google Drive, among others), and all
simulators of one tenant are projections of a single underlying entity
graph. The account that appears as a CRM opportunity is the same
account referenced by a support ticket and a recorded call, under one
join key, so questions that span systems are well defined by
construction. Agents access the environment through each simulator's
MCP tools, and the data plane is read-only, which prevents an agent
under evaluation from modifying the data against which it is graded.

The estate also contains databases for the fictional company's own
applications and analytics. No vendor fixes their schemas. The
benchmark therefore uses a specialized question-conditioned generator
for these databases. A benchmark author supplies business questions, a
language model designs the required schema once, and deterministic
code creates the records for each seed. The completed entity graph is
an input to this process. Shared entities and attributes are copied
from the graph. The generator creates only the facts specific to the
internal application. A consistency check rejects any database whose
shared keys or values disagree with the graph. The two mechanisms
therefore populate the same company.

The benchmark computes every expected answer from the records an agent
can access. Questions over product simulators come from code templates
that are instantiated with the generated company's entities. Questions
over internal databases guide schema and record generation. The
generator plants qualifying rows and near misses, then computes the
final answer from all completed rows. Both mechanisms also support
questions whose correct answer is that the requested information is
absent.

The two generation mechanisms require different checks
(Section~\ref{sec:fidelity}). Internal databases have explicit column
plans and planting directives. Their checks measure plan adherence,
question specificity, label integrity, and dependency structure. The
entity graph has no paired real company for comparison. Its realism is
scored against targets from Eon's operational data and published
sources, and an adversarial detector searches for synthetic artifacts.
The scorecard and detector guided revisions of the entity-graph
generator. Across 23 generated companies, the mean realism score rose
from 61.8 to 97.0, and the share of records flagged as synthetic fell
from 55.2\% to zero.

The contributions of this paper are: (1) a generator that produces a
complete fictional estate spanning vendor applications and internal
databases; (2) a specialized generator that turns internal-database
questions into schemas, planted records, and exact answer keys; (3)
validation methods scoped to each generation mechanism; and (4) a
deterministic evaluation of nine models over 33 questions, with three
repeats and paired statistical tests.

\section{Related work}
\label{sec:related}

\textbf{Enterprise agent benchmarks.}
$\tau$-bench~\citep{yao2024tau} grades agents on the final state of a
hand-written database and introduces the pass\textasciicircum{}$k$
metric for consistency across repeated trials; its authors report that
most failures are errors of database reasoning rather than of
conversation. CRMArena~\citep{huang2025crmarena} populates a Salesforce
organization from a hand-designed schema with latent variables, and
generates its task instances with an LLM followed by human
verification. CRMArena-Pro~\citep{huang2025crmarenapro} extends the
design to multi-turn interactions and confidentiality-awareness tests.
TheAgentCompany~\citep{xu2024theagentcompany} measures completion of
long-horizon work tasks in a self-hosted office environment, where the
best agent completes roughly a quarter of the tasks.
WorkArena~\citep{drouin2024workarena} and
WorkArena++~\citep{boisvert2024workarenapp} evaluate browser agents on
ServiceNow with task templates over the product's demo data. The benchmark
differs from all of these in three respects: the environment spans
many systems that share one entity graph rather than one system; the
answers are computed from the generated records rather than maintained
by hand; and the realism of the data is measured rather than assumed.
It also generates the schemas and records of company-specific
databases from author-supplied business questions.

\textbf{Text-to-SQL benchmarks.} BIRD~\citep{li2023bird} showed that
much of the difficulty of answering questions over real databases
comes from the data itself, including inconsistent and dirty values,
and includes questions that the database cannot answer, graded on
abstention. Spider~2.0~\citep{lei2025spider2} moved evaluation to
enterprise warehouses, where schema size and dialect differences
dominate. The benchmark adopts both lessons at generation time: controlled doses
of dirty values are injected into the generated records, and labeled
unanswerable questions are part of every benchmark.

\textbf{Synthetic data fidelity.} The Synthetic Data
Vault~\citep{patki2016sdv} and its SDMetrics library define the
standard metrics for synthetic data, all of which compare against the
real dataset being imitated. Hudovernik et
al.~\citep{hudovernik2024syntherela} show that detection by a strong
classifier is the most discriminating evaluation for relational data,
and that child tables are synthesized worse than parent tables.
Zhang~\citep{zhang2026dependency} shows that the commonly used linear
classifier two-sample test~\citep{lopezpaz2017c2st} is nearly blind to
dependencies between columns, since independently shuffled data passes
it; the boosted-tree variant does not share this blindness.
TabQueryBench~\citep{liu2026tabquerybench} evaluates synthetic data by
the analytical queries it can answer correctly and reports that the
best generators recover only about 40\% of rare values.
Hollywood~\citep{stoian2026hollywood} generates a synthetic film
database and argues its realism from the cardinality-estimation
errors~\citep{moerkotte2009qerror} it induces in a query optimizer,
which are comparable to those induced by the real database.
Synthea~\citep{walonoski2018synthea} generates synthetic patient
records from published clinical statistics, and is the closest
precedent for the benchmark's approach of building realism from cited
reference statistics rather than from training data. It operationalizes
these results for a setting none of them addresses: benchmark data with
no paired real counterpart.

\section{One company, many systems}
\label{sec:estate}

\subsection{The entity graph}

The benchmark represents the company facts shared by product
simulators and internal databases in one entity graph. One seeded
generator produces its identity (name and mail domain, derived from
the chosen industry and business model), its workforce, its customers,
and the business activity that connects them.

The workforce is generated first. The composition of the workforce
follows the company's size and growth stage: a two-hundred-person
company at one stage is a fifth sales and a third engineering, and the
proportions shift with the stage. Each employee has a position on a
job ladder, a manager, a hire date drawn from realistic tenure
distributions, a country with its own holidays and working hours, and
an employment type (employee or contractor). Some employees have
already left the company; the records they created remain, owned by a
person who is no longer present. This is generated deliberately,
because access-review and offboarding questions are a recurring
category of enterprise work.

The customer base is generated next and assigned to sales
representatives as territories of roughly twenty accounts each, with a
cap that prevents any single representative from holding a
disproportionate share. Under a business-to-consumer configuration a
customer is an individual rather than a company, which changes what
every downstream system holds. Around the customer base the generator
produces activity in dependency order: contacts, sales deals with full
stage history, documents, chat channels and messages, support tickets,
recorded calls, and marketing campaigns. Later records reference
earlier ones, so a ticket's requester is a contact of a real account
and the owner of a deal is a member of the sales team.

Volumes and values are correlated rather than drawn independently.
Larger accounts produce more tickets. Meetings fall on business days
in the participants' time zones, with a small realistic share of
weekend exceptions. Timestamps respect the life cycle they belong to:
a deal cannot close before it opens, and a record that is still open
has no end date. The graph also carries the texture that distinguishes
an operating company from a demonstration dataset: duplicate customer
records, a large body of inactive prospect accounts alongside the real
customer base, non-human service accounts, a knowledge base derived
from the company's own support tickets, and the company's own
infrastructure with services and incidents.

Generation is deterministic. Each subsystem of the generator draws
from its own seeded random stream, so the same seed reproduces the
identical company on any machine, and extending one subsystem does not
perturb the output of another. Determinism is what allows every
simulator, every evaluation run, and every regression test to observe
exactly the same company.

\subsection{Projection into product simulators}

Each business application in the environment is a simulator: a service
that implements the interface of a real product. The fleet contains
simulators of 66 products, including Salesforce, HubSpot, Zendesk,
Jira, Gong, Slack, Stripe, S3, and Google Drive; a provisioned
environment enables a configured subset. A simulator does not generate
data of its own. It rebuilds the tenant's entity graph from the same
seed and configuration and projects it: it stores the slice of the
graph its product would hold, in that product's own format. No data is
transferred between simulators; determinism is the synchronization
mechanism.

Projections are faithful to each vendor and deliberately partial. The
same recorded call appears in the Gong simulator as a call with
participants and a transcript, and in the Salesforce simulator as a
logged activity attached to the same contact and deal. Duplicate
customer records appear only in the CRM, since a typographical
duplicate is a CRM phenomenon. Prospect lists appear only in systems
that hold an account table. Documents are rendered once as real bytes
(valid DOCX, XLSX, and PDF files whose contents agree with the tabular
records that reference them) and distributed to the file-store
simulators. Record identifiers are derived deterministically in each
vendor's real format, so vendor client libraries validate them, and
each identifier maps back to the shared entity, which is what keeps a
cross-system join well defined.

\subsection{The application portfolio}

A tenant declares which applications the company runs; we call this
set the portfolio. The graph is always generated in full, regardless
of the portfolio, so two tenants that differ only in portfolio agree
record for record on everything both include. The portfolio is
consulted only at projection time, under two rules. First, a simulator
whose product is not in the portfolio serves a valid empty workspace
rather than an error. Second, the traces that one product's
integration writes into another exist only when the counterpart is in
the portfolio. In a tenant that runs Gong, call records in Salesforce
carry the artifacts the real Gong integration writes, including
recording links; in a tenant without Gong, the same calls occur at the
same times with the same participants, but no Gong artifact appears
anywhere. The underlying business is thus independent of the software
choices, which keeps results comparable across portfolio
configurations. This distinction is not cosmetic: in production
Salesforce organizations that run Gong, a substantial fraction of the
activity table is written by the Gong integration package.

\subsection{Serving the estate}

A provisioned environment, called an estate, is created from the four
parameters of Section 1. The provisioning service builds every
selected simulator, waits until all report ready, and returns one MCP
endpoint per simulator together with a single tenant-bound credential.
The credential's tenant is a signed claim, so one tenant cannot read
another's data. The data plane is read-only: a request that would
modify data is refused, whether it arrives through a vendor API or an
MCP tool, and the refusal itself is useful in evaluation, since an
agent that attempts a prohibited write fails the test rather than the
deployment. The environment evolves through day states: day 0 is the
empty system, day 1 the full history, and day 2 adds one further day
of change, with the answer key recomputed at each state, so staleness
and incremental-synchronization behavior are themselves measurable.

\section{Questions with computed answers}
\label{sec:questions}

\subsection{Simulator question and answer generation}

Grading an agent requires a set of questions and the correct answer to
each. In prior benchmarks both are authored manually, and the answers
must be maintained by hand against the data. For product-simulator
scenarios, the benchmark derives both from the generated company itself.

The benchmark includes 83 scenarios: one for each combination of
ten industries and six size tiers, and 23 individually named
companies. A scenario is a generated company together with its
questions and their computed answers, all created in advance and
committed to the corpus. A continuous-integration test regenerates the
corpus from its seeds and verifies that the result is byte-identical
to the committed files, so the benchmark cannot drift as the generator
evolves.

The questions in every scenario come from a shared battery of
templates written in code. Each template specifies a question form
together with the computation that answers it: the number of open
sales opportunities, the total value of the open pipeline, the
salesperson with the highest won revenue, or the median resolution
time of a support ticket. When a scenario is built, each template is
instantiated with that company's entities, so the question text refers
to specific accounts and channels of that company, and the expected
answer is computed directly from the generated records. Because the
answer is derived from the data, it cannot disagree with the data.

This design makes extending the benchmark inexpensive. Adding a
scenario requires only a new company specification, namely an
industry, a size tier, and a seed; the templates instantiate against
the new company and the answers are computed automatically. Adding a
question requires only a new template, which then applies to every
scenario. No answer is ever written by hand.

Each scenario also stores consistency checks. A consistency check
states that a named entity, such as a specific account, must appear in
several named systems and refer to the same entity in all of them. A
verifier reads those systems and confirms it. Cross-system questions
depend on this agreement, so it is checked rather than assumed.

\subsection{Verifying that answers are reachable}

An answer computed from the graph is correct by construction, but
correctness alone is insufficient: the underlying fact must also be
reachable through the product interfaces an agent uses. A separate
audit therefore re-reads every audited fact through each simulator's
native API and through its MCP tools, and asserts that the value read
back equals the stored answer. This audit has identified real defects.
An early cross-system question joined recorded calls to contacts
through an internal record identifier; the join was correct against
the stored data but impossible through the product interface, because
the call system issues its own participant identifiers and only the
email address is visible on both sides. The audit rejects such
questions before they can penalize a correct agent.

\subsection{Composed questions and abstention}

Single-system questions measure retrieval. The benchmark also includes
twelve composed questions that require work across systems. They are
defined the same way as the questions above, as code that pairs a
question text with the computation of its answer. A composed question
spans several systems and several steps: totals grouped by an attribute
held in another system, the three accounts with the most support
tickets, the recorded call minutes of the account with the largest
open pipeline, or the number of accounts that raised an urgent ticket
after their first open opportunity. The same twelve questions apply to
every scenario, and each answer is recomputed from that scenario's
records.

Three rules keep these questions fair. First, every answer computation
joins systems only through keys an agent can follow in the products
themselves; for example, the call system does not expose CRM
identifiers, so calls are joined to accounts through participant email
addresses. Second, each question states its own definitions and
tie-breaking rule in its text, so a model is never graded against an
unstated convention. Third, when a scenario's data does not settle a
question uniquely, for instance when two accounts tie for the largest
open pipeline although the question implies a single winner, the
question is excluded for that scenario rather than graded against an
arbitrary choice.

Two of the twelve questions request information the environment does
not contain, such as a Net Promoter Score no system stores, and the
correct response is to state that the question cannot be answered.
Grading is deterministic in every case: each question declares its
comparison mode, namely an exact value, a set, an ordered sequence, a
per-key mapping, or abstention.

\subsection{Generating internal databases from question lists}
\label{sec:synthgen}

The same fictional estate also contains databases of its own, such as
an analytics warehouse or an application database. Their schemas are
particular to the company and fixed by no vendor. The benchmark uses a
specialized generator to create these internal databases. It is an
internal development tool: clients receive the finished estate and
answer keys, not the generator itself. Its input is a list of business
questions about the fictional company and the completed entity graph.
The internal databases extend that graph with application-specific
facts. Their customers, employees, and business events reuse the
graph's canonical entities.

The design phase happens once per benchmark, and it is the only phase
in which a language model participates. A benchmark author writes a
list of business questions in natural language. The model also receives
a catalog of the entity types, identifiers, and attributes available
in the graph. It first sketches an application that could answer the
questions: which databases exist, which entities live in each, and how
they relate. It then produces the concrete schema, a generation plan
for every column, and a planting directive for every question. Each
table and column is marked as graph-backed or database-specific. A
graph-backed record must use the graph's canonical identifier. Any
repeated name, date, amount, or other shared attribute must also come
from the graph.

The model also proposes a few questions that the database cannot
answer, whose purpose is to test whether an agent admits when an answer
does not exist. For each such question, the model must state which
table or column the database would need in order to answer it. A
validator checks the schema, including every mapping to the entity
graph. It rejects a design that creates a second identity for a shared
entity. It also rejects a proposed unanswerable question when the
schema contains the required information. All model output is
validated, mechanically repaired where the mistake is clerical, and
scored. A weak design buys a bounded number of redesign turns, and the
best valid design is kept. The result is a fixed design.

The generation phase is deterministic and runs after the entity graph
has been completed. It first materializes graph-backed records and
foreign keys from the graph. It then generates database-specific
tables and columns in dependency order from the plan's typed
distributions. A repeated account, employee, contact, deal, or event is
copied from the graph rather than drawn again. New internal records
reference those shared entities through their canonical identifiers.

The generator then plants each question. It forces a stated number of
rows to satisfy the question's conditions, so no question is
accidentally empty or trivial. Each planted question also receives
near-miss rows. Every near miss satisfies all but one condition, with
the violated condition rotating across rows. A solver that ignores any
single condition therefore reaches a wrong answer. Planting may add
database-specific facts to a graph entity, but it cannot alter a
graph-backed identifier or shared attribute.

Controlled doses of dirty values, such as inconsistent casing and
placeholder text, are applied last, following the observation of
BIRD~\citep{li2023bird} and Spider~2.0~\citep{lei2025spider2} that much
of the difficulty of real data lies in the data itself. Graph-backed
keys and values are protected from these changes. A final consistency
gate resolves every shared reference to the graph and compares every
copied value with its source. The database is rejected if an identity
is missing, a foreign key does not resolve, or a shared value differs.
Because the product simulators project the same graph, this gate also
keeps the internal databases consistent with the simulator records.

The answer key is computed at the very end, by running every question
against the finished tables. The planted rows only ensure that matching
rows exist. The recorded answer counts all rows that satisfy the
question, including any that match by coincidence. The answer
therefore always agrees with the data a solver actually reads. Each
stored answer carries its comparison mode and, for fractional values,
a numeric tolerance.

The same seed reproduces the identical database byte for byte, and a
different seed keeps the schema and structure while changing the
values. Every database generated from one design therefore carries the
same questions, each with answers recomputed from its own records,
exactly as the scenario templates above are fixed in form while their
answers are computed per company.

\section{Validating the generated estate}
\label{sec:fidelity}

\subsection{Checks follow the generation mechanism}

Established methods for evaluating synthetic data assume that a real
dataset exists for comparison: the generator is trained on real
records, and quality is measured as the statistical distance between
the synthetic output and that original~\citep{patki2016sdv}. The Era
by Eon Benchmark generates fictional companies without paired real
records. Distance-based evaluation therefore does not apply.

Each generation mechanism provides different evidence. Internal
databases have a stored schema, a generation plan for every column, and
planting directives for their questions. They also record the entity
graph source of every shared table, key, and value.
Section~\ref{sec:refree} checks the finished tables and labels against
that declared design and against the graph. Product simulators derive
from the same entity graph. Their records are audited through the
interfaces available to an agent, and the graph is scored against
realism targets from Eon's operational data and published statistics.
The provenance of each target is recorded, and estimates are marked.

\subsection{Checks for the internal databases}
\label{sec:refree}

An internal database can be defective even when its design is valid.
The generator can fail to produce the declared data. A question can be
answerable by shortcut. Its answer key can be ambiguous, or declared
relationships can be absent from the rows. The benchmark first runs an
estate-consistency gate. Every graph-backed identifier must resolve to
the canonical entity, and every copied value must equal its graph
source. Since each simulator is a projection of the graph, this also
checks consistency with the records exposed through product
interfaces.

The benchmark then runs four measurements over every internal
database. Each compares the finished database with its stored design
or with a transformed copy of its rows, so none requires external data.

The first measurement, adherence, verifies that the rows deliver what
the design declares. The generation plan of every column states its
domain: the permitted categories of a status field, the minimum and
maximum of a numeric field, the fraction of values that may be null.
After generation, the actual rows are compared against these
declarations. A status value that is not among the declared
categories, an amount outside its declared range, or a column declared
five percent null that comes out thirty percent null are all
violations. Such a violation matters beyond aesthetics: a question
that filters on the affected column no longer means what the design
says it means, so its planted answer may be wrong. Because the design
itself passed validation, any disagreement between design and rows is
a bug in the generation code, and this measurement is the only place
where that class of bug is visible.

The second measurement asks whether each question actually requires
the work it appears to require. Consider a question with three
conditions, such as jobs that failed, in one region, and are not
deleted. If nearly every failed job is in that region and not deleted,
then a solver that checks only the first condition reaches the same
answer as one that checks all three, and the question rewards
carelessness as much as care. To detect this, every question is
answered by a panel of deliberately lazy solvers: one drops a
condition, one ignores the join between tables and reads a single
table, one aggregates over all rows instead of the filtered ones. Any
question that a lazy solver answers exactly correctly is measuring
nothing, and it is reported so that its data can be re-planted or the
question replaced. The near-miss rows of
Section~\ref{sec:synthgen} exist precisely to make these lazy
strategies fail.

The third measurement examines the answer key itself. Three defects
make a key unfair rather than a question hard. First, a tie: if the
question asks for the three accounts with the most tickets and the
third and fourth place hold the same count, two different answers are
equally correct, and whichever the key recorded, a solver giving the
other is marked wrong. Second, self-answering text: a question whose
own wording contains its answer is a copying exercise. Third,
precision: a fractional answer such as an average must carry an
explicit tolerance, because a solver's own arithmetic legitimately
differs from the key's in the last decimal places, and exact string
comparison would mark it wrong. All three defects are searched for
mechanically, and a benchmark is not released while any is present.

The fourth measurement tests whether the dependencies the design
declares between columns exist in the delivered rows. This is the
failure that standard single-column statistics cannot see. Take any
table and shuffle each column independently: every single-column
distribution is preserved perfectly, yet every relationship between
columns is destroyed. A dataset and its shuffled copy are
indistinguishable to marginal tests, and prior work has shown that
even the widely used linear classifier two-sample
test~\citep{lopezpaz2017c2st} barely separates
them~\citep{zhang2026dependency}. The benchmark therefore uses the stronger form
of the test: a boosted-tree classifier is trained to distinguish the
generated rows from an independently shuffled copy of themselves. If
the classifier succeeds, the rows carry structure beyond their
marginals, which is what the design promised; if it performs at
chance, the declared dependencies were never generated. The
classifier's score is read against a measured noise floor, obtained by
scoring shuffled data against itself, so that a table with no declared
dependencies scores zero rather than a spurious positive value. A
complementary check computes the cardinality-estimation
error~\citep{moerkotte2009qerror} that the data induces in a textbook
query optimizer: real data, with its correlations and skew,
systematically misleads optimizers that assume independence and
uniformity, and Hollywood~\citep{stoian2026hollywood} argues that
synthetic data is realistic to the extent that it misleads them
comparably. This error is computable with no real data at all.

\subsection{The realism scorecard}
\label{sec:scorecard}

The realism scorecard evaluates the entity graph and its product
projections. It asks whether the shared company records resemble an
operating company. A graph can be internally consistent while every
histogram is flat, every account holds the same number of contacts, and
every timestamp advances in fixed steps. The scorecard compares each
generated company with the reference targets described above, along
five axes. Each axis addresses one way in which generated data
typically betrays itself.

The marginal axis scores the distribution of each field on its own.
Naive generation draws values uniformly, which produces perfectly flat
histograms; in real systems, ticket priorities follow a skewed mix in
which urgent tickets are rare, and deal sizes follow a heavy-tailed
distribution in which a few large deals coexist with many small ones.
Each field's empirical distribution is tested against its reference
target.

The joint axis scores whether fields that should move together
actually do. In a real company, a customer's renewal risk reflects its
open ticket load, revenue grows with headcount as a scattered cloud
rather than an exact line, and larger accounts spend more. Generated
data whose columns are drawn independently fails all of these, and an
analytical question over such data, for instance which account is most
at risk, has an answer that is statistical noise. The axis tests the
declared relationships with rank correlations between the fields
involved.

The temporal axis scores the calendar. Real business activity
concentrates in working hours in each participant's own time zone,
thins at weekends, and spikes at quarter ends when sales teams pull
deals to the boundary. Events also follow their causes: a call about a
deal happens after the deal opens, and a ticket's resolution follows
its creation by a duration that depends on its priority. The axis
tests both the time-of-day and calendar structure and the causal
ordering.

The structural axis scores the shape of relationships. In a real
customer base, a small number of accounts carry a large share of the
tickets and the pipeline, and a long tail of accounts is nearly
inactive. Generated data that gives every account the same number of
contacts, deals, and tickets is recognizable at a glance. The axis
tests per-parent counts for the heavy-tailed concentration the
references specify.

The content axis scores the text. Ticket subjects, ticket bodies, chat
messages, call transcripts, and deal notes must be varied at corpus
scale, and each record must be about one thing: a ticket whose subject
reports a permissions failure while its body discusses a scheduled
export is nonsense to a human reader even when every aggregate over it
is correct. The axis measures vocabulary growth against the sublinear
curve a human-written corpus of the same size would show, and checks
that all text on one record describes the same issue.

Each axis produces a score between 0 and 100. Two safeguards protect
the meaning of these scores. The first concerns small companies. A
statistical check needs a minimum amount of data: a distribution
cannot be judged from a dozen values, and a concentration curve cannot
be judged from four accounts. When a generated company is too small
for a check, that check is skipped and reported as not measured,
instead of contributing a meaningless pass or fail. Every scorecard
also reports what fraction of its checks actually ran, so a company
that scored well on the few checks it was large enough to support
cannot be mistaken for one that passed a full examination. The second
safeguard concerns overfitting by the developers of the generator. All
generation is seeded, and during development the generator is tuned
while observing scores on particular seeds. If the scores were checked
only on those same seeds, a change could raise them in ways that hold
for those seeds alone. The automated test suite therefore scores every
change on a fixed seed that was never used during development, and
rejects the change if the scores regress there.

\subsection{The adversarial detector and the measured improvement}

The scorecard measures how well the data matches reference statistics.
A second instrument examines the data the way a suspicious human
reader would: it looks for the specific marks that give synthetic data
away. These marks are well known. A generator that runs out of names
adds a number to make new ones, so a customer list reads Dunn, Dunn2,
Dunn3. A field whose values are drawn uniformly produces a histogram
flatter than any real system shows. Timestamps produced by a counter
advance in identical steps. Every parent record has exactly the same
number of children. The same filler sentence appears on hundreds of
records. The detector scans every generated record for these marks and
reports the fraction of records that carry at least one. Unlike the
scorecard, it does not reward the data for matching anything; it
actively hunts for evidence of generation.

The scorecard and the detector were built first, and the entity-graph
generator was improved afterward, guided by what they reported. This
order gives a clean before-and-after measurement. The original
entity-graph generator, scored over 23 generated companies, had a mean
realism score of 61.8, its worst company scored 48.6, and the detector
flagged 55.2\% of all records. The generator was then revised in a
series of passes. Each pass fixed what the instruments reported as
worst at that point:
uniform draws were replaced with the reference distributions,
per-account volumes were made heavy-tailed, a business calendar was
introduced so events fall at realistic times and in causal order,
fields that should be correlated were coupled, all text on a record
was made to describe one issue, and numbered name variants were
replaced with plausible names. After the fifth pass, the mean score
over the same 23 companies was 97.0, the worst company scored 92.8,
and the detector flagged no records. Later passes raised the mean to
98.6.

During this work, some checks in the scorecard were found to be
flawed. They gave low scores to small companies even when the data was
realistic, because the statistic they computed requires more data than
a small company has. For example, one check measured the variety of
the written text and expected a large vocabulary; a company with only
three support tickets has too little text to reach it, no matter how
well written. Another check measured the percentage of events that
fall within business hours from only sixteen timestamps; with a sample
that small, the percentage swings widely by chance alone. These flawed
checks were fixed so that they take into account how much data they
are given. After the fixes, every score in this section, including the
original 61.8 baseline, was recomputed with the final version of the
scorecard. The improvement from 61.8 to 97.0 is therefore a comparison
of two generators under one identical scorecard, not an artifact of
changing the scoring rules midway.

\section{Evaluating agents on the estate}
\label{sec:eval}

\subsection{The evaluation harness}

The environment described so far exists to answer one question: how
well does an agent work over a company's business systems? The
evaluation harness measures this directly. It takes one committed
scenario, provisions the simulators with that scenario's records, asks
a model the scenario's questions, and grades every answer against the
stored answer key.

The model reaches the data the same way a deployed agent would,
through MCP. Each simulator exposes its product's operations as MCP
tools; a tool is one callable operation, such as running a query in
the CRM, listing the tickets of the support desk, or fetching the
transcript of a recorded call. Each simulator exposes tens of such
tools, and across the simulators of one scenario the model faces
roughly 290 of them in total. Nothing about this surface is described
in the model's prompt: no schemas, no data description, no tool list.
The model starts with only three meta-tools, one that lists the
systems present in the environment, one that describes the tools of a
chosen system, and one that calls a chosen tool. It must therefore
discover for itself which systems exist, which tools they offer, and
which of them answers the question at hand. This discovery is part of
what the benchmark measures, while the size of the prompt is not,
because the tool catalog never enters it.

Four controls keep the comparison between models fair. First,
isolation: every run is provisioned with its own copy of the scenario,
torn down when the run ends, so concurrent runs cannot see each
other's data. Second, read-only access: only tools that cannot modify
data are exposed, so no run can change the records that a later
question is graded against. Third, equal budgets: the reported
comparison allows 25 turns, 40 tool calls, and 8{,}000 characters per
tool response. Every run records whether it answered, exhausted a
budget, or failed to produce a well-formed answer. Fourth,
verified serving: before any question is asked, the harness reads each
simulator back and checks that it serves exactly the scenario's
records. If a simulator fails this check, its questions are reported
as not asked instead of being counted as model failures, so a defect
in our serving cannot be mistaken for a model error.

Grading is deterministic. Each stored answer declares its comparison
mode, as described in Section~\ref{sec:questions}, and the grader
applies that mode mechanically. The reported comparison gives no
partial credit and uses no model as a judge. Each repeat receives a
fresh isolated tenant seeded from the same frozen scenario.

\subsection{Full model comparison}

We evaluated nine models on the same frozen
\texttt{segment\_fintech\_small} estate. Each model answered 33
questions three times, giving 99 recorded answers per model and 891
answers in total. The set contains 21 single-system questions and 12
composed questions. It uses Gong, Salesforce, and Zendesk records.
The source corpus contained 67 candidate questions. Eight were omitted
because they required unsupported HubSpot, Jira, Slack, or cross-system
identity operations. Another 26 concerned documents and other
unstructured artifacts that were not exposed in this tool-use track.
These exclusions were fixed before any model ran and did not affect the
scores. Every model received the same remaining 33 questions.

Accuracy is first averaged across the three repeats of each question
and then across the 33 questions. Table~\ref{tab:model-results} reports
these estimates. A correct abstention counts as correct. A false
abstention means that the model refused an answerable question.
``No answer'' combines malformed outputs and runs that ended before an
answer. Each outcome count is out of 99.

\FloatBarrier
\begin{table}[ht]
\caption{Nine-model comparison on one generated estate. Each 95\%
interval shows the uncertainty caused by evaluating only 33 questions
with three runs per question. FA denotes false abstention; NA denotes
no answer.}
\label{tab:model-results}
\centering
\footnotesize
\begin{tabular}{lrrrrr}
\toprule
Model & Accuracy & 95\% interval & Wrong & FA & NA \\
\midrule
\texttt{claude-opus-4.8}   & 76.8\% & [61.6, 90.9] &  0 & 10 & 13 \\
\texttt{gpt-5.6-sol}       & 69.7\% & [54.5, 83.8] & 17 & 13 &  0 \\
\texttt{claude-opus-5}     & 67.7\% & [52.5, 81.8] &  1 & 10 & 21 \\
\texttt{claude-sonnet-5}   & 67.7\% & [51.5, 82.8] &  4 &  7 & 21 \\
\texttt{gpt-5.5}           & 66.7\% & [50.5, 81.8] &  0 & 17 & 16 \\
\texttt{claude-sonnet-4.6} & 63.6\% & [47.5, 78.8] & 16 & 13 &  7 \\
\texttt{gpt-5.6-luna}      & 56.6\% & [40.4, 72.7] & 26 & 12 &  5 \\
\texttt{qwen3-coder-next}  & 55.6\% & [39.4, 71.7] & 14 &  3 & 27 \\
\texttt{gpt-5.4-mini}      & 42.4\% & [27.3, 58.6] & 41 & 14 &  2 \\
\bottomrule
\end{tabular}
\end{table}
\FloatBarrier

The benchmark contains only 33 questions, so small score differences
may result from the choice of questions. The intervals in the table
show this uncertainty. Wide intervals mean that the exact ordering of
the models is uncertain.

We also checked every pair of models to determine whether one performed
reliably better than the other. Only three of the 36 comparisons
provided strong enough evidence. The remaining score differences do
not support a claim that one model is better.

Three differences passed this test: Claude Opus 4.8 exceeded GPT-5.4
Mini by 34.3 percentage points; GPT-5.6 Sol exceeded GPT-5.4 Mini by
27.3 points; and Claude Sonnet 4.6 exceeded GPT-5.4 Mini by 21.2
points. Every other pairwise result was inconclusive.

The final results show a consistent pattern of difficult work. Three
questions appeared among the five hardest for all nine models: finding
the longest Gong call, totaling all Gong call time, and reporting the
ticket-priority distribution for the account with the most tickets.
For seven of the nine models, the five hardest questions included one
that combines Salesforce and Gong records. It first asks the agent to
identify the customer account with the largest total value of open
sales opportunities. It then asks for the total recorded call time for
that account. No model answered the median ticket-resolution-time
question correctly.

The capability scores show the same pattern. Averaged across the nine
models, accuracy was 3.7\% on multi-hop questions, 6.2\% on
superlatives, 7.4\% on temporal questions, 11.1\% on ranking and
order-sensitive questions, and 39.6\% on cross-system questions.
Simple filtering reached 92.6\%. The main difficulty therefore lies in
exhaustive aggregation, multi-hop joins, temporal summaries, and
ordered rankings rather than basic filtering.

The 891 runs produced 561 correct answers, 119 wrong answers, 99 false
abstentions, six malformed outputs, and 106 runs that ended before an
answer. List-price inference for these 891 recorded runs totaled
\$343.27.

\section{Limitations}
\label{sec:limitations}

The model comparison covers one generated company and 33 independent
question shapes. Three repeats reduce model-generation noise but do
not create more questions. The benchmark contains separate questions
for internal databases and documents. The reported comparison evaluates
only the 33 simulator questions that use Gong, Salesforce, and Zendesk.
It therefore provides no results for the other existing questions or
for the full corpus of 83 scenarios.

The realism targets are only as good as their sources. Most are drawn
from Eon's operational data or from published statistics, but some are
estimates, and although every estimate is marked as such, a wrong
target pulls the generator toward a wrong distribution. The scorecard
measures agreement with the targets, not the targets' own truth.

The generated text is produced from seeded templates with
slot-filling. The content axis and the adversarial detector verify
variety and record-level coherence, and the current corpus passes
both, but template-generated text still has a ceiling: it cannot reach
the full diversity, idiosyncrasy, and error patterns of human writing,
and a stronger detector than ours might separate the two.

\section{Conclusion}

The Era by Eon Benchmark generates the entire evaluation environment
for enterprise LLM agents: one fictional company, the vendor
applications and internal databases it operates, the benchmark
questions, and the exact answer to each question. Two generation
mechanisms populate this estate. The entity graph supplies shared
facts to the product simulators. The specialized internal-database
generator turns business questions into schemas and records. It copies
shared entities, identifiers, and values from the graph before adding
application-specific facts. The simulators and internal databases
therefore describe the same company. Every answer is computed from the
final records, so grading is mechanical.

Generating everything creates an obligation that hand-built benchmarks
do not face: the realism of the data must be demonstrated rather than
assumed. Each mechanism supplies the evidence appropriate to its
output. Design-based checks verify the internal databases, their
questions, and their labels. The entity-graph scorecard grades
generated companies against statistics of real business data, and the
adversarial detector hunts for synthetic artifacts. The scorecard and
detector drove the entity-graph generator's mean realism score across
23 companies from 61.8 to 97.0 and reduced the flagged share from
55.2\% to zero.

The full comparison contains 891 recorded answers from nine models.
Accuracy estimates range from 42.4\% to 76.8\%. The broad intervals
and corrected paired tests show that most apparent differences between
models remain inconclusive. The benchmark therefore reports both point
estimates and statistical evidence. Its purpose is to measure factual
tool use over a coherent company with exact ground truth.

\bibliographystyle{plainnat}
\bibliography{era}

\end{document}